\def\year{2019}\relax
\documentclass[letterpaper]{article} 
\usepackage{aaai19}  
\usepackage{times}  
\usepackage{helvet}  
\usepackage{courier}  
\usepackage{url}  
\usepackage{graphicx}  
\usepackage{amsfonts}       
\usepackage{times}
\usepackage{latexsym}
\usepackage{graphicx,float,array,multirow}
\usepackage{url}
\usepackage{amsmath}
\usepackage{siunitx,caption}
\usepackage{subcaption}
\usepackage[linesnumbered,ruled]{algorithm2e}
\usepackage{mathtools}
\usepackage{booktabs}
\usepackage{nicefrac}       
\usepackage{microtype}      

\DeclareMathOperator*{\argmax}{argmax}

\begin{document}
%
\title{Adversarial Dropout for Recurrent Neural Networks}
\author{Sungrae Park$^{1}$\footnote{This work was done at KAIST.}
    \hspace{0.15in} Kyungwoo Song$^{2}$ 
    \hspace{0.15in} Mingi Ji$^{2}$
    \hspace{0.15in} Wonsung Lee$^{3}$\footnotemark[\value{footnote}] 
    \hspace{0.15in} Il-Chul Moon$^{2}$ \\
$^{1}$ Clova AI Research, NAVER Corp., Korea  \\
$^{2}$ Industrial \& Systems Engineering, KAIST, Korea \\
$^{3}$ AI Center, SK Telecom, Korea \\
}

\maketitle
\begin{abstract}
Successful application processing sequential data, such as text and speech, requires an improved generalization performance of recurrent neural networks (RNNs). Dropout techniques for RNNs were introduced to respond to these demands, but we conjecture that the dropout on RNNs could have been improved by adopting the \textit{adversarial} concept. This paper investigates ways to improve the dropout for RNNs by utilizing intentionally generated dropout masks. Specifically, the guided dropout used in this research is called as \emph{adversarial dropout}, which adversarially disconnects neurons that are dominantly used to predict correct targets over time. Our analysis showed that our regularizer, which consists of a gap between the original and the reconfigured RNNs, was the upper bound of the gap between the training and the inference phases of the random dropout. We demonstrated that minimizing our regularizer improved the effectiveness of the dropout for RNNs on sequential MNIST tasks, semi-supervised text classification tasks, and language modeling tasks.
\end{abstract}

\section{Introduction}
\label{introduction}
 
Many effective regularization methods have been introduced to address the issue of large-scale neural networks predisposed to overfitting. Among several regularization techniques \cite{bishop1995training,ioffe2015batch,salimans2016weight}, dropout \cite{srivastava2014dropout,kingma2015variational} has become a common methodology because of its simplicity and effectiveness. The dropout randomly disconnects neural units during training to prevent the feature co-adaptation. Srivastava et al. \cite{srivastava2014dropout} interpreted the dropout as an extreme form of a model ensemble by sharing the extensive parameters of a neural network. However, the naive application of the dropout to hidden states of recurrent neural networks (RNNs) failed to prove performance gains \cite{zaremba2014recurrent} because it interferes with abstracting long-term information. This issue is caused by the structural differences between feed-forward neural networks (FFNNs) and RNNs. 

Recent works have investigated the applications of the dropout on the recurrent connections of RNNs. Gal and Ghahramani \cite{gal2016theoretically} suggested the application of the dropout on the same neural units through time steps, as opposed to dropping different neural units at each time step. Their approach is similar to the L2 regularization on the weight parameters in the recurrent connection. After this suggestion, Semeninuta et al. \cite{Semeniuta45423} and Moon et al.\cite{moon2015rnndrop} explored the dropout applications within long-short term memory (LSTM) cells by regulating the gating mechanism. Recently, Merity et al. \cite{merity2017regularizing} proposed the use of \emph{DropConnect} \cite{wan2013regularization} on the recurrent hidden-to-hidden weight matrices. This approach also allows the recurrent units to share the same dropout mask. These dropout techniques for RNNs have shown that recurrent units should have the same transition metric to process the sequential information. These results are used in extending the dropout techniques on RNNs. 

Recently, dropout-based ensemble (DE) regularization \cite{bachman2014learning,ma2016dropout,Laine2017iclr,tarvainen2017mean} has been established to improve the dropout techniques. DE regularization conceptually consists of two phases: the generation of the dropout masks and the comparison of the original and the perturbed networks by the dropout masks. Bachman et al. \cite{bachman2014learning} and Ma et al. \cite{ma2016dropout} tried to minimize the distance between the output distributions of the original network and its randomly perturbed network. Laine and Aila \cite{Laine2017iclr} suggested self-ensembling models, or the $\Pi$ model, containing a distance term between two randomly perturbed networks. 
Park et al. \cite{park2017adversarial} proposed \emph{adversarial dropout} that intentionally deactivates neurons that are dominantly used to predict the correct target. By utilizing the adversarial dropout, the regularization consists of the distance between the original and the adversarially perturbed networks. Additionally, they proved further improvements for supervised and semi-supervised image classification tasks. These DE regularizers are fundamentally developed for all types of neural networks, but these DE regularizations are not compatible to RNNs because of the sequential information abstraction. 

To investigate the effectiveness of the DE regularization on RNNs, Zolna et al. \cite{zolna2018fraternal} proposed \emph{fraternal dropout} (FD), an RNN version of \emph{self-ensembling} \cite{Laine2017iclr}. Specifically, two randomly perturbed RNNs by \emph{DropConnect} are simultaneously supervised from the true labels while maintaining the distance between their predictions. They showed further generalization improvements on the language modeling, considered to be one of the most demanding applications of RNNs. Their results motivated us to extend the \emph{adversarial dropout} to RNNs.

This paper proposes \textit{adversarial dropout} on recurrent connections to adaptively regularize RNNs. Similar to the adversarial dropout on FFNNs, adversarial dropout for RNNs also deactivates dominating hidden states to predict the correct target sequence. In order to adapt the concept of the adversarial dropout to the RNN structure, first, we analytically showed that the regularizer, which consists of a gap between the original and the perturbed RNNs, is the upper bound of the gap between the training and the inference phases of the random dropout on an RNN \cite{bachman2014learning,ma2016dropout}. Second, we investigated ways to measure which recurrent neurons are mainly used to predict target sequences in using a time-invariant dropout mask. Finally, we improved the algorithm to find a better adversarial dropout condition compared to the previous algorithm finding the adversasrial dropout. 
According to our experiments, adversarial dropout for RNNs showed the advanced performances on the sequential versions of MNIST, the semi-supervised text classification, and the language modeling tasks. 

\section{Preliminary}
\subsection{Recurrent neural networks with dropout noise}
\begin{figure}
\centering
  	\includegraphics[width=0.45 \textwidth]{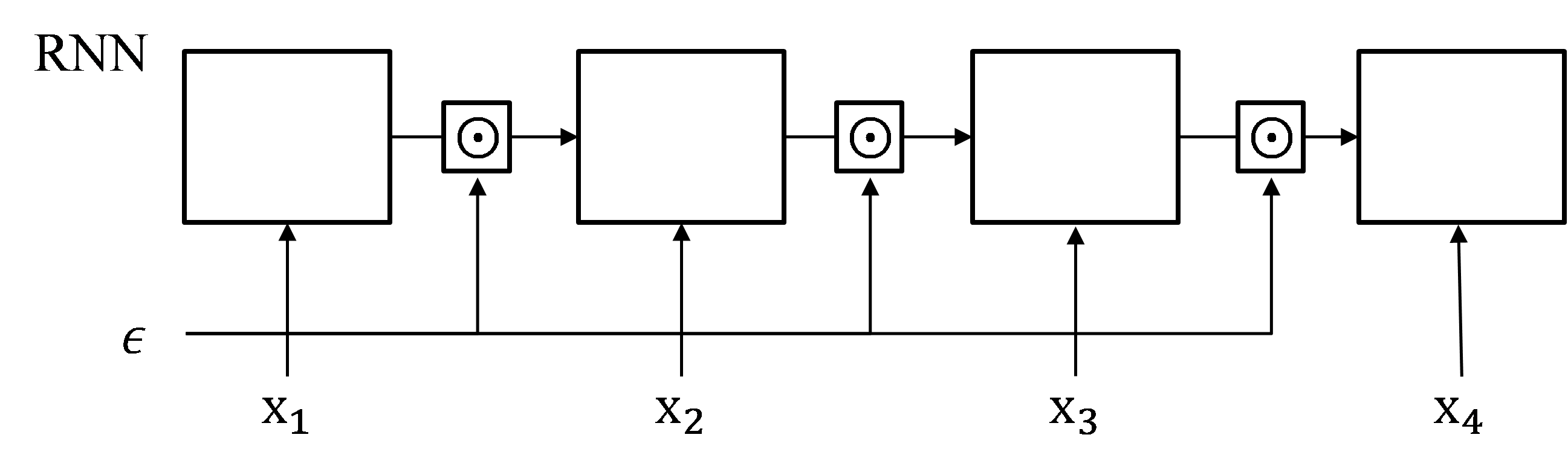}
  \caption{A recurrent neural network with dropout noise on recurrent connections. Each recurrent connection shares the time-invariant dropout mask, $\epsilon$, to alleviate the destructive loss.}
\end{figure}
We denoted simple RNN models for brevity of notation because derivations for LSTM and GRU follow similarly. Given input sequence $\boldsymbol{\mathrm{x}}=\{\mathrm{x}_1,..., \mathrm{x}_T\}$ of length $T$, a simple RNN model is formed by the repeated application of function $f_h$. This generates a hidden state $\boldsymbol{h}_t$ at time step $t$ by taking $\text{x}_t$ and $\boldsymbol{h}_{t-1}$ as input. For descriptions of the dropout for RNNs, we denoted $d( \boldsymbol{h}, \boldsymbol{\epsilon} ) =  \boldsymbol{h} \odot \boldsymbol{\epsilon} / (1-p) $ as the random dropout function where $\boldsymbol{h}$ is the dropout target and $\boldsymbol{\epsilon}$  is the dropout mask, sampled from the Bernoulli distribution with success probability $(1-p)$. Gal et al. \cite{gal2016theoretically} proposed variational dropout to drop the previous hidden state $\boldsymbol{h}_{t-1}$ with the same dropout mask for every time steps. The following is the RNN equation with the variational dropout:
\begin{equation}
\boldsymbol{h}_t = f_h(\mathrm{x}_t, \boldsymbol{h}_{t-1}) = \sigma(\mathrm{x}_t \boldsymbol{W}_h + d( \boldsymbol{h}_{t-1}, \boldsymbol{\epsilon} ) \boldsymbol{U}_h + \boldsymbol{b}_h ) 
\end{equation}
where $\boldsymbol{W}_h$ and $\boldsymbol{U}_h$ are parameter matrices that model input-to-hidden and hidden-to-hidden (recurrent) connections, respectively; $\boldsymbol{b}_h$ is a vector of bias terms; and $\sigma$ is the nonlinear activation function. We notice that all hidden states share the time-invariant dropout mask, $\boldsymbol{\epsilon}$. Figure 1 describes the dropout noise space on the recurrent connections sharing the same mask across every time steps. The goal of this paper is to investigate the effectiveness of adversarial dropout on RNNs. When we applied the adversarial dropout perturbation to the model, we replaced the dropout mask, $\boldsymbol{\epsilon}$, with adversarial dropout mask, $\boldsymbol{\epsilon}^{\text{adv}}$.

\subsection{Inference gap of dropout for recurrent neural networks}

Recent results have shown that there is a gap between objectives for training and testing when using a dropout technique. This is called the \emph{inference gap of dropout}\cite{ma2016dropout,bulo2016dropout}. Dropout training indicates learning an ensemble of neural networks, but the output of each network in the ensemble should be averaged to provide the final prediction. Unfortunately, this averaging over an exponential number of subnetworks is intractable, and standard dropout typically implements an approximation by applying the expectation of the dropout variable to compute outputs with a deterministic network. However, the approximation causes the gap between the training and the inference phase of the dropout techniques. 

Expectation-linearization (EL) regularization \cite{ma2016dropout}, which is similar to a pseudo-ensemble agreement \cite{bachman2014learning}, tries to reduce the gap by adding a penalty to the objective function. Let $\boldsymbol{\epsilon}$  be the dropout mask applied to the recurrent connection of RNNs, and $\boldsymbol{p}_t(\boldsymbol{\mathrm{x}}, \boldsymbol{\epsilon} ; \boldsymbol{\theta})$ be the prediction of the model for the input sequence $\boldsymbol{\mathrm{x}}$ at time $t$. The EL regularizer can be represented as a generalized form as shown below:
\begin{equation}
\mathcal{R}^{\text{EL}}(\boldsymbol{\mathrm{x}};\theta) :=  \mathrm{E}_{\boldsymbol{\epsilon}}\Big[ \sum_{t=1}^{T} \lambda_t \mathrm{D}\big[\boldsymbol{p}_t(\boldsymbol{\mathrm{x}},\mathrm{E}_{\boldsymbol{\epsilon}}[\boldsymbol{\epsilon}] ; \boldsymbol{\theta} ) || \boldsymbol{p}_t(\boldsymbol{\mathrm{x}},\boldsymbol{\epsilon} ; \boldsymbol{\theta} ) \big] \Big]
\end{equation}
where $\mathrm{D}[\cdot || \cdot]$ indicates a non-negative function that represents the distance between two output vectors such as cross-entropy, $\boldsymbol{\epsilon}$ is a Bernoulli random vector with the probability $p$, and $\lambda_t$ is a hyperparameter controlling the intensity of the loss at time $t$. The goal of the EL regularizer is to minimize network loss as well as the expected difference between the prediction from the random dropout mask and the prediction from the expected dropout mask, which is fully connected. However, the EL regularizer requires a Monte Carlo (MC) approximation because calculating the penalty is still intractable due to the expectation. 

FD regularization \cite{zolna2018fraternal}, which is an RNN version of $\Pi$ models \cite{Laine2017iclr}, indirectly reduces the inference gap by minimizing the variance of output distributions caused by random dropout masks. Its regularizer consists of the distance between two outputs perturbed by two sampled dropout masks as shown in the following:
\begin{equation}
\mathcal{R}^{\text{FD}}(\boldsymbol{\mathrm{x}};\theta) :=  \mathrm{E}_{\boldsymbol{\epsilon}^1, \boldsymbol{\epsilon}^2}\Big[ \sum_{t=1}^{T} \lambda_t \mathrm{D}\big[\boldsymbol{p}_t(\boldsymbol{\mathrm{x}}, \boldsymbol{\epsilon}^1 ; \boldsymbol{\theta} ) || \boldsymbol{p}_t(\boldsymbol{\mathrm{x}},\boldsymbol{\epsilon}^2 ; \boldsymbol{\theta} ) \big] \Big],
\end{equation}
where $\boldsymbol{\epsilon}^1$ and $\boldsymbol{\epsilon}^2$ are Bernoulli random vectors with the probability $p$. Zolna et al. \cite{zolna2018fraternal} proved that the FD regularizer is related to the lower bound of the EL regularizer, $\frac{1}{4}\mathcal{R}^{\text{FD}}(\boldsymbol{x};\theta) \leq  \mathcal{R}^{\text{EL}}(\boldsymbol{x} ;\theta)$. To apply the FD regularizer, the MC approximation is required due to the expectation over the dropout masks.

We should note that the adversarial dropout for RNNs consists of the distance between the output distributions perturbed by a base dropout mask and an adversarial dropout mask, $\boldsymbol{\epsilon}^{\text{adv}}$. In section 3, we prove that our regularizer is the upper bound of the EL regularizer without any further MC approximation.

\subsection{Adversarial training}

\emph{Adversarial training} is a training process that uses adversarial examples crafted to include a small perturbation in inputs so as to not classify the perturbed inputs correctly \cite{goodfellow2014explaining,papernot2016limitations,miyato2017virtual}. The perturbation is a real-valued vector and is mainly used to generate intentional noisy data. Recently, the concepts of adversarial training has been extended to sequential modeling tasks. Samanta et al. \cite{samanta2017towards} proposed an algorithm that crafts adversarial text examples by using gradient-based saliency maps. They proved that the retraining with the crafted text improves generalization performance on the text classification task. In addition to directly crafting the adversarial text examples, Miyato et al. \cite{miyato2016adversarial} injected adversarial noise in the word-embedding layer and built a regularizer including adversarial perturbation. They evaluated adversarial training on semi-supervised text classification tasks and they achieved state-of-the-art performances. In contrast to the adversarial perturbation on input space, adversarial dropout identifies subnetworks misclassifying data even though a few neurons are deactivated. The successful applications of adversarial training on RNNs become additional motivation to extend adversarial dropout to RNNs. Additionally, our results showed that the model, when combining adversarial training and adversarial dropout, achieved an advanced level of  performance on Internet Movie Database (IMDB). 

\section{Adversarial dropout for recurrent neural networks}

\emph{Adversarial dropout} \cite{park2017adversarial} is a novel DE regularization to reduce the distance between two output distributions of the original network and an intentionally perturbed network. Specifically, the adversarial dropout generate a perturbed network by disconnecting neural units that are dominantly used to predict correct targets. 
The perturbed network reconfigured by the adversarial dropout might not classify the correct target even though few neurons are deactivated from the full network. Therefore, its learning stimulates the useless or the incorrectly learned neurons to better contribute to more accurate predictions. The following is an RNN version of the adversarial dropout regularizer:

\begin{gather}
\label{eq_adv}
 \mathcal{R}^{\text{AdD}}(\boldsymbol{\mathrm{x}}, \boldsymbol{\epsilon}^{0} ;\theta) := \sum_{t=1}^{T} \lambda_t \mathrm{D}\big[\boldsymbol{p}_t(\boldsymbol{\mathrm{x}},\boldsymbol{\epsilon}^{0} ; \boldsymbol{\theta} ) || \boldsymbol{p}_t(\boldsymbol{\mathrm{x}},\boldsymbol{\epsilon}^{\text{adv}} ; \boldsymbol{\theta} ) \big] \\[-1ex]
\text{where} \:\: \boldsymbol{\epsilon}^{\text{adv}}:=\argmax_{{\boldsymbol{\epsilon}} ; \| \boldsymbol{\epsilon} - \boldsymbol{\epsilon}^0 \|_2 \leq \delta} \sum_{t=1}^{T} \lambda_t  \mathrm{D}\big[\boldsymbol{p}_t(\boldsymbol{\mathrm{x}},\boldsymbol{\epsilon}^{0} ; \boldsymbol{\theta} ) || \boldsymbol{p}_t(\boldsymbol{\mathrm{x}},\boldsymbol{\epsilon} ; \boldsymbol{\theta} ) \big].
\end{gather} 

\noindent where $\delta$ is the hyperparameter controlling the intensity of the noise. 
In this equation, $\boldsymbol{\epsilon}^{0}$ is the base dropout mask, which represents a target network that supervises an adversarially dropped network. For a example, $\boldsymbol{\epsilon}^{0}$ can be set as $\boldsymbol{1}$ vector that indicates the original network without any dropped neurons. At each training step, we identified the worst case dropout condition, $\boldsymbol{\epsilon}^{\text{adv}}$, against the current model, $\boldsymbol{p}_t(\boldsymbol{x},\boldsymbol{\epsilon}^{0}; \boldsymbol{\theta} )$, and trained the model to be robust to such dropout perturbations by minimizing the regularization term. 

We note that the regularization term is equivalent to three statistical relationships between two output distributions by a dropout mask, $\boldsymbol{\epsilon}^0$, and its adversarial dropout mask, $\boldsymbol{\epsilon}^{\text{adv}}$, as shown below (proof in the appendix).

\textbf{Remark 1.}  Let $\boldsymbol{\epsilon}$ be an i.i.d. dropout mask; 
$p_{t, i}(\boldsymbol{\epsilon})$ be a $i^{\text{th}}$ element of $\boldsymbol{p}_t(\boldsymbol{x},\boldsymbol{\epsilon} ; \boldsymbol{\theta} ) \in \mathbb{R}^M$ where $M$ is the size of the output dimension; 
$\mathcal{R}_t^{\text{AdD}}(\boldsymbol{x}, \boldsymbol{\epsilon} ;\theta)$ be the distance term at time $t$ in Eq.4; and $\epsilon^{\text{adv}}$ be an adversarial dropout mask by setting the base dropout mask as $\boldsymbol{\epsilon}$ when the distance metric is L2 norm. Then, 
{\setlength{\abovedisplayskip}{-1pt}
\setlength{\belowdisplayskip}{-0.5pt}
\begin{gather}
\begin{split}
\mathrm{E}_{\boldsymbol{\epsilon}}[\mathcal{R}_t^{\text{AdD}}(\boldsymbol{x}, \boldsymbol{\epsilon} ;\theta)] = 
& \overbrace{ \sum_i \mathrm{V}_{\boldsymbol{\epsilon}}[p_{t, i}(\boldsymbol{\epsilon})] + \sum_i \mathrm{V}_{\boldsymbol{\epsilon}}[p_{t, i}(\boldsymbol{\epsilon}^{\text{adv}})] }^{\text{(1)}}  \\[-2ex]
& -2 \overbrace{ \sum_i \mathrm{Cov}_{\boldsymbol{\epsilon}}[p_{t, i}(\boldsymbol{\epsilon}), p_{t, i}(\boldsymbol{\epsilon}^{\text{adv}})]}^{\text{(2)}} \\[-2ex]
& + \overbrace{ \sum_i \big(\mathrm{E}_{\boldsymbol{\epsilon}}[p_{t, i}(\boldsymbol{\epsilon})]- \mathrm{E}_{\boldsymbol{\epsilon}}[p_{t, i}(\boldsymbol{\epsilon}^{\text{adv}})]\big)^2}^{\text{(3)}} ,
\end{split}
\end{gather} }

where $\mathrm{V}_{\boldsymbol{\epsilon}}$ is the variance and $\mathrm{Cov}_{\boldsymbol{\epsilon}}$ is the covariance. We note that $\boldsymbol{\epsilon}^{\text{adv}}$ depends on the base dropout mask, $ \boldsymbol{\epsilon}$, so its output, $\boldsymbol{p}_t(\boldsymbol{x},\boldsymbol{\epsilon}^{\text{adv}} ; \boldsymbol{\theta} )$, contains randomness derived from $\boldsymbol{\epsilon}$. Minimizing the regularization in Eq.5 pursues (1) invariant outputs over random dropout masks and adversarial dropout masks, (2) positive relationship between the two outputs, and (3) minimized distance between the means of the two outputs. The first terms (1) are consistent with the goal of the FD regularization that minimizes the variance of the output distributions over dropout masks \cite{zolna2018fraternal}. The second term (2) indicates that outputs move in the same direction even though dominantly used neurons for the correct predictions are deactivated in the adversarially perturbed network. If the output of sub-network A is higher than the output of sub-network B, the output of adversarial sub-network from A becomes higher than the one from B.  This means that a paired gap between a sub-network and its adversarial sub-network becomes similar to the other paired gaps caused by dropping dominantly used neurons. This interpretation provides the connection between one paired case to the others. The third term (3) minimizes the mean of the gaps. This might increase training error of the base network because the adversarially perturbed networks have poor performance. On the other hand, it can improve generalization performance by preventing the base network from overfitting to training data. Figure 2 shows a graphical description of the three statistical relationships. 
\begin{figure}
\centering
  	\includegraphics[width=0.45 \textwidth]{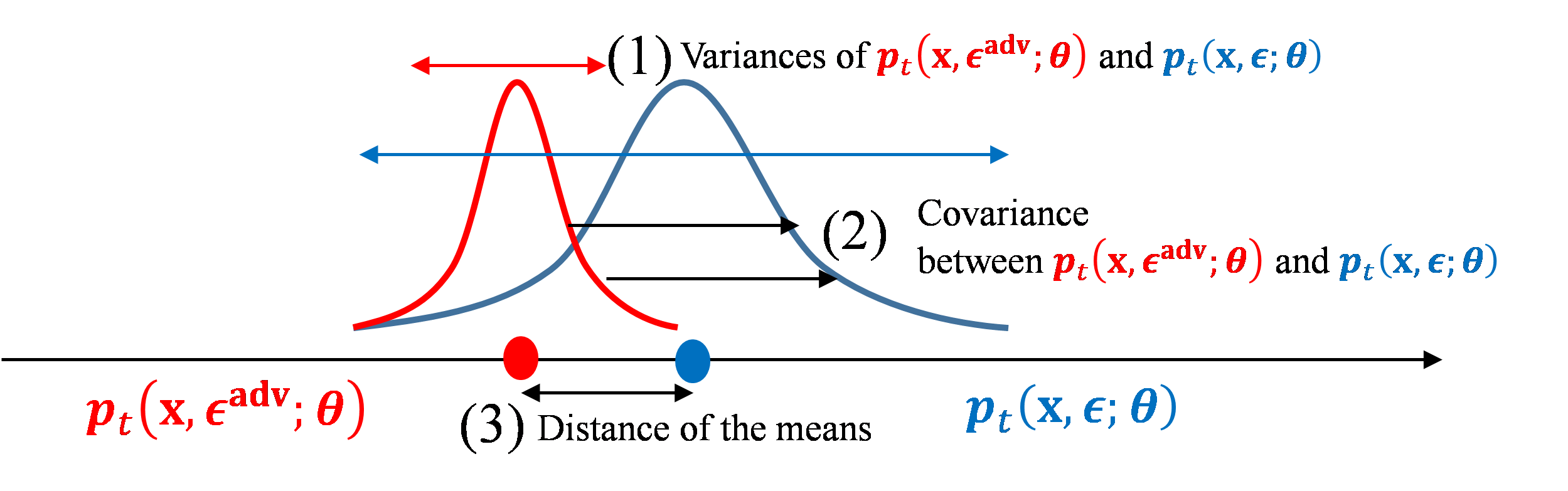}
  \caption{Regularization term of adversarial dropout indicates three statistical relationships between the outputs of randomly dropped networks and of adversarially dropped networks: (1) the variances of two outputs, (2) the covariance between two outputs, and (3) the distance between the means.}
\belowdisplayskip=-2pt
\end{figure}

When comparing our regularizer with the other EL regularizer, our regularization is the upper bound of the EL regularization when $\boldsymbol{\epsilon}^0=\mathrm{E}_{\boldsymbol{\epsilon}}[\boldsymbol{\epsilon}]$ without the MC approximation under the constrained domain space of the dropout masks, $\boldsymbol{\epsilon}$ and $\boldsymbol{\epsilon}^0$ (proof in the appendix). 

\noindent \textbf{Proposition 1.}  $\frac{1}{4}\mathcal{R}^{\text{FD}}(\boldsymbol{x};\theta) \leq \mathcal{R}^{\text{EL}}(\boldsymbol{x};\theta) \leq  \mathcal{R}^{\text{AdD}}(\boldsymbol{x}, \mathrm{E}_{\boldsymbol{\epsilon}}[\boldsymbol{\epsilon}] ;\theta)$ when $\boldsymbol{\epsilon} \in \{ \boldsymbol{\epsilon} | \| E_{\boldsymbol{\epsilon}}[\boldsymbol{\epsilon}] - \boldsymbol{\epsilon} \|_2 \leq \delta\} $.

\noindent We note that our regularization is also positioned in the upper bound of the FD regularization. This result shows that our regularization has relevance with others. The adversarial dropout has a largest upper bound, and this suggests the robustness, yet slow learning. Given a set of regularizers, one cannot tell which one is a right structure for his/her domain. Often, researchers just try multiple cases and choose one experimentally. 

\subsection{Finding adversarial dropout masks for recurrent neural networks}

In order to identify $\boldsymbol{\epsilon}^{\text{adv}}$, we needed to minimize the distance term in Eq.4 with respect to $\boldsymbol{\epsilon}$. However, exact minimization is intractable for modern neural networks because of nonlinear activations of the network and the discrete domain of $\boldsymbol{\epsilon}$. This challenge led us to approximate the worst case of the dropout masks by applying the following two steps: (1) identifying the influences of the recurrent neurons and (2) applying a greedy algorithm based on the influence scores \cite{park2017adversarial}. 

The influences of the recurrent neurons should be time-invariant because the recurrent neurons are repeatedly applied over every time step. Fortunately, we can derive the influences by using a gradient of the distance term with respect to the dropout mask because the recurrent neurons have a time-invariant dropout mask. To get the influences, we first relax the discrete domain of  $\boldsymbol{\epsilon} \in \{0, 1\}^D$ to the continuous domain, $\tilde{\boldsymbol{\epsilon}} \in [0, 1]^D$, then we calculate the gradients with respect to $\tilde{\boldsymbol{\epsilon}}$. We named the gradient values as influence map (IM) because they indicated the influences of the recurrent neurons. 

Let $\boldsymbol{h}_t(\tilde{\boldsymbol{\epsilon}})$ and $\hat{\boldsymbol{h}}_t(\tilde{\boldsymbol{\epsilon}})$ be a hidden state and a dropped hidden state by $\tilde{\boldsymbol{\epsilon}}$ at time $t$, and $\mathrm{D}_t(\tilde{\boldsymbol{\epsilon}})=\mathrm{D}[\boldsymbol{p}_t(\boldsymbol{x}, \boldsymbol{\epsilon}^0 ; \boldsymbol{\theta} ) || \boldsymbol{p}_t(\boldsymbol{x},\tilde{\boldsymbol{\epsilon}} ; \boldsymbol{\theta} )]$. The IM can be calculated as follows:
\begin{gather}
\begin{split}
\mathrm{IM}_i (\boldsymbol{\epsilon}^{s}) &:= \frac{\partial \sum_{t=1}^{T} \lambda_t \mathrm{D}_t(\tilde{\boldsymbol{\epsilon}}) }{\partial \tilde{\epsilon}_i}\Big|_{\tilde{\boldsymbol{\epsilon}}=\boldsymbol{\epsilon}^s} \\
&=  \sum_{t=1}^{T} \lambda_t \frac{\partial \mathrm{D}_t(\boldsymbol{\epsilon}^s) }{\partial \hat{\boldsymbol{h}}_{t}(\boldsymbol{\epsilon}^s)} \Big\{ \sum_{u=1}^{t-1} \frac{\partial \hat{\boldsymbol{h}}_{t}(\boldsymbol{\epsilon}^s)}{\partial \hat{h}_{u, i}(\boldsymbol{\epsilon}^s)} h_{u, i}(\boldsymbol{\epsilon}^s) \Big\}, 
\end{split}\end{gather}
where $\boldsymbol{\epsilon}^s$ is the initial dropout mask of $\tilde{\boldsymbol{\epsilon}}$. In the equation, $h_{u, i}(\tilde{\boldsymbol{\epsilon}})$ and $\hat{h}_{u, i}(\tilde{\boldsymbol{\epsilon}})$ are the $i^{\text{th}}$ components of $\boldsymbol{h}_t(\tilde{\boldsymbol{\epsilon}})$ and $\hat{\boldsymbol{h}}_t(\tilde{\boldsymbol{\epsilon}})$ respectively. The equation on the right shows an alternative view of the IM without any Jacobian matrices with respect to $\tilde{\boldsymbol{\epsilon}}$. The equation contains the gradient propagation, $\partial \hat{\boldsymbol{h}}_{t}(\boldsymbol{\epsilon}^s)) / \partial \hat{h}_{u, i}(\boldsymbol{\epsilon}^s))$, which is calculated by the backpropagation of the recurrent directions from time $u$ to time $t$. This alternative equation shows that the IM provide the degree of the influence of the recurrent units over time. Additionally, the IM depends on the initial values of the dropout mask, $\boldsymbol{\epsilon}^s$. It is important to note that the initial dropout mask, $\boldsymbol{\epsilon}^s$, should be different from $\boldsymbol{\epsilon}^0$ because the same conditions of dropout mask cause the zero values of the gradient if there is no another stochastic process in the model. In this paper, we initialized $\boldsymbol{\epsilon}^s$ by randomly flipping a element of the base dropout mask $\boldsymbol{\epsilon}^0$ to set a initial dropout mask $\boldsymbol{\epsilon}^s$. Calculated IM values indicate the relations between the recurrent neurons and the distance term. 

After calculating the IM, the adversarial dropout mask, $\boldsymbol{\epsilon}^{\text{adv}}$, should be identified under the constraint, $\| \boldsymbol{\epsilon}^{\text{adv}} - \boldsymbol{\epsilon}^0 \|_2 \leq \delta$. For the worst case dropout conditions, the neuron with a positive IM value should be activated and the neuron with a negative IM value should be deactivated. Park et al. \cite{park2017adversarial} proposed a greedy algorithm to find the worst case of the dropout mask utilizing the IM. The algorithm estimates IM values once and iteratively flips a dropout element shown to be highly effective to the distance term in view of the IM. The adversarial dropout with the greedy algorithm is shown below:
\begin{equation}
\boldsymbol{\epsilon}^{\text{adv}} = Flip_{\| \boldsymbol{\epsilon}^{s} - \boldsymbol{\epsilon}^0 \|_2 \leq \delta} \big( \boldsymbol{\epsilon}^{s}, \mathrm{IM}(\boldsymbol{\epsilon}^{s}) \big),
\end{equation}
where we introduce a flip function, which changes elements of the dropout mask, $\boldsymbol{\epsilon}^{s}$, in the descending order of $(1-2\boldsymbol{\epsilon}^{s})\odot\mathrm{IM}(\boldsymbol{\epsilon}^{s})$ until the constraint $\| \boldsymbol{\epsilon}^{s} - \boldsymbol{\epsilon}^0 \|_2 \leq \delta$ is satisfied (detail algorithm in appendix). In the flip function, $(1-2\boldsymbol{\epsilon}^{s})\odot\mathrm{IM}(\boldsymbol{\epsilon}^{s})$ indicates the influence of flipping a dropout element because the relation estimated by the IM depends on the state of the current dropout mask, $\boldsymbol{\epsilon^{s}}$. This approach is simple but can be improved because IM values are unstable when an element is flipped. 

We constructed a straightforward way to extend the greedy algorithm. Specifically, we modified the greedy algorithm to update IM values multiple times, as shown here: 
\begin{gather} \begin{split}
&\boldsymbol{\epsilon}^{\text{adv}, (0)} = \boldsymbol{\epsilon}^{s}, \\
&\boldsymbol{\epsilon}^{\text{adv}, (k+1)} =
Flip_{\| \boldsymbol{\epsilon}^{\text{adv}, (k+1)} - \boldsymbol{\epsilon}^{0} \|_2 \leq \frac{k+1}{K}\delta} \big( \boldsymbol{\epsilon}^{\text{adv}, (k)}, \mathrm{IM}(\boldsymbol{\epsilon}^{\text{adv}, (k)}) \big),
\end{split}\end{gather}
where $K$ is the maximum number of the iteration. For a better approximation of the worst-case dropout mask, the IM values are updated using the current dropout mask and the bound of the constraint is increased through the iterations. There is a trade-off between the better approximation and the computational cost. The large number of the iteration, $K$, leads to a better approximation, but it also increases the computational cost proportionately. In our experiment, we tested cases of $K$=1 and $K$=2 to investigate performance improvement obtained by sacrificing the computational cost. 

\section{Experiments}
\subsection{Sequential MNIST}

Sequential MNIST tasks, also known as pixel-by-pixel MNIST, process each image one pixel at a time and predicts the label of the image. The tasks can be categorized by the order of the pixels in a sequence: $s$MNIST in scanline order and $p$MNIST in a fixed random order. Because the size of an MNIST image is 28 $\times$ 28, the length of the sequence becomes 784.

Our baseline model included a single LSTM layer of 100 units with a softmax classifier to produce a prediction from the final hidden state. For the settings for the dropout, we set the dropout probability as 0.1 for the baseline models. In the case of the adversarial dropout, we adapted $\boldsymbol{\epsilon^{0}}=\mathrm{E}_{\boldsymbol{\epsilon}}[\boldsymbol{\epsilon}]$, which indicates the expectation of the dropout mask, and $\delta=0.03$, which represents the maximum changes from the base dropout mask as 3\%. These hyperparameters of the baseline models as well as our models were retrieved in the validation phase. All models were trained with the same optimizer (detail settings in appendix). 

\begin{table}
\centering
\caption{Test error rates of supervised learning on $s$MNIST and $p$MNIST. Each setting is repeated ten times.}
\begin{tabular}{lcc}
\toprule
  Model  & sMNIST & pMNIST\\
\midrule
Unregularized  					&1.018 ($\pm$0.165)		&9.924 ($\pm$0.307) \\   
VD \shortcite{gal2016theoretically}		&0.721 ($\pm$0.111)		&5.218 ($\pm$0.132)   \\
FD \shortcite{zolna2018fraternal}		&0.720 ($\pm$0.061)		&5.121 ($\pm$0.121) \\
AD ($K$=1)     					&0.705 ($\pm$0.047) &5.046 ($\pm$0.067) \\
AD ($K$=2)     					&\textbf{0.644} ($\pm$0.046)&\textbf{5.030} ($\pm$0.065) \\
\hline
\end{tabular}
\end{table}

\begin{figure}[t]
\centering
\includegraphics[width=\linewidth]{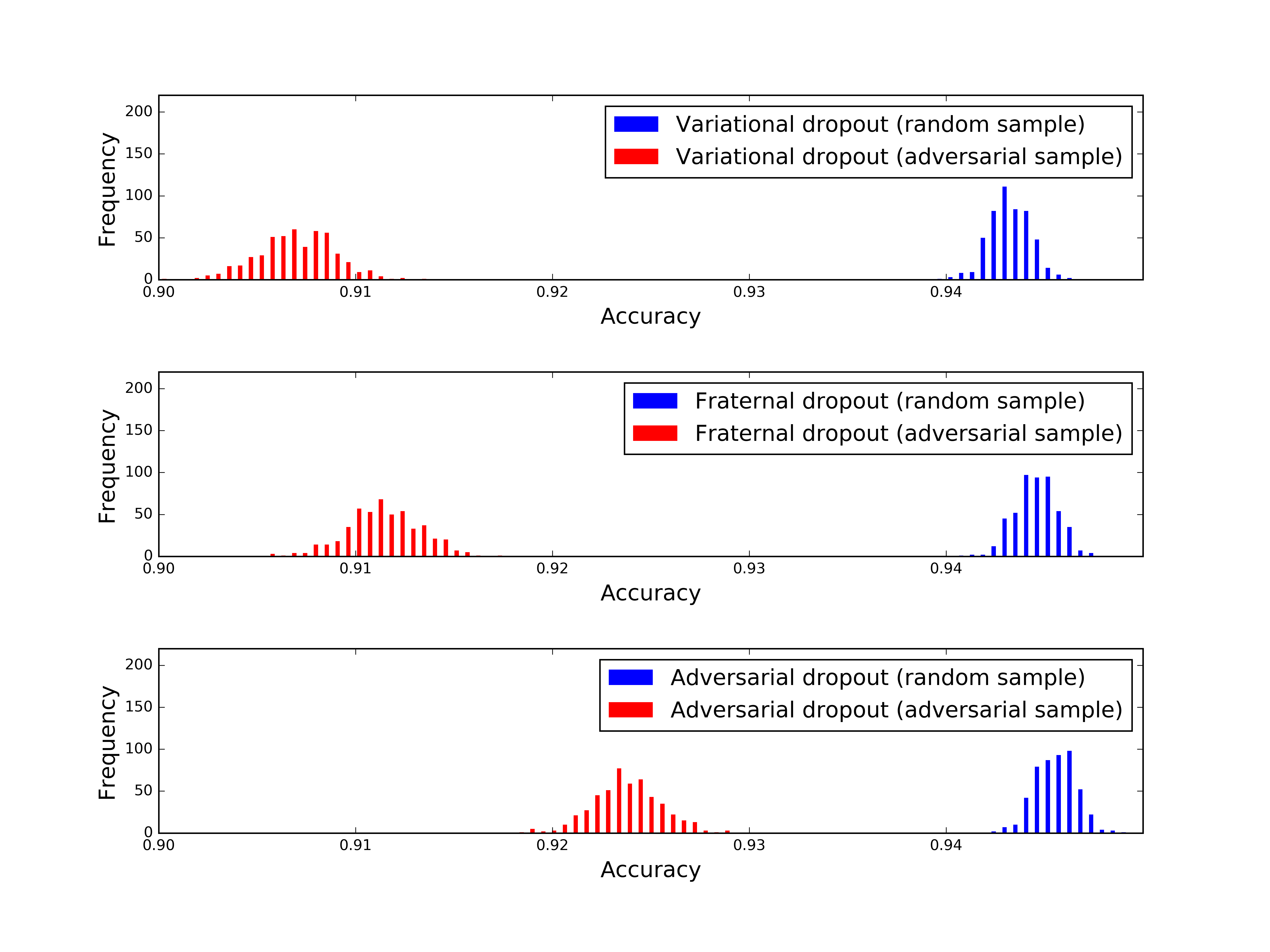}
\caption{Test accuracies of the perturbed RNNs by random dropout and adversarial dropout masks.}
\end{figure}

\begin{figure}[t]
\centering
\includegraphics[width=\linewidth]{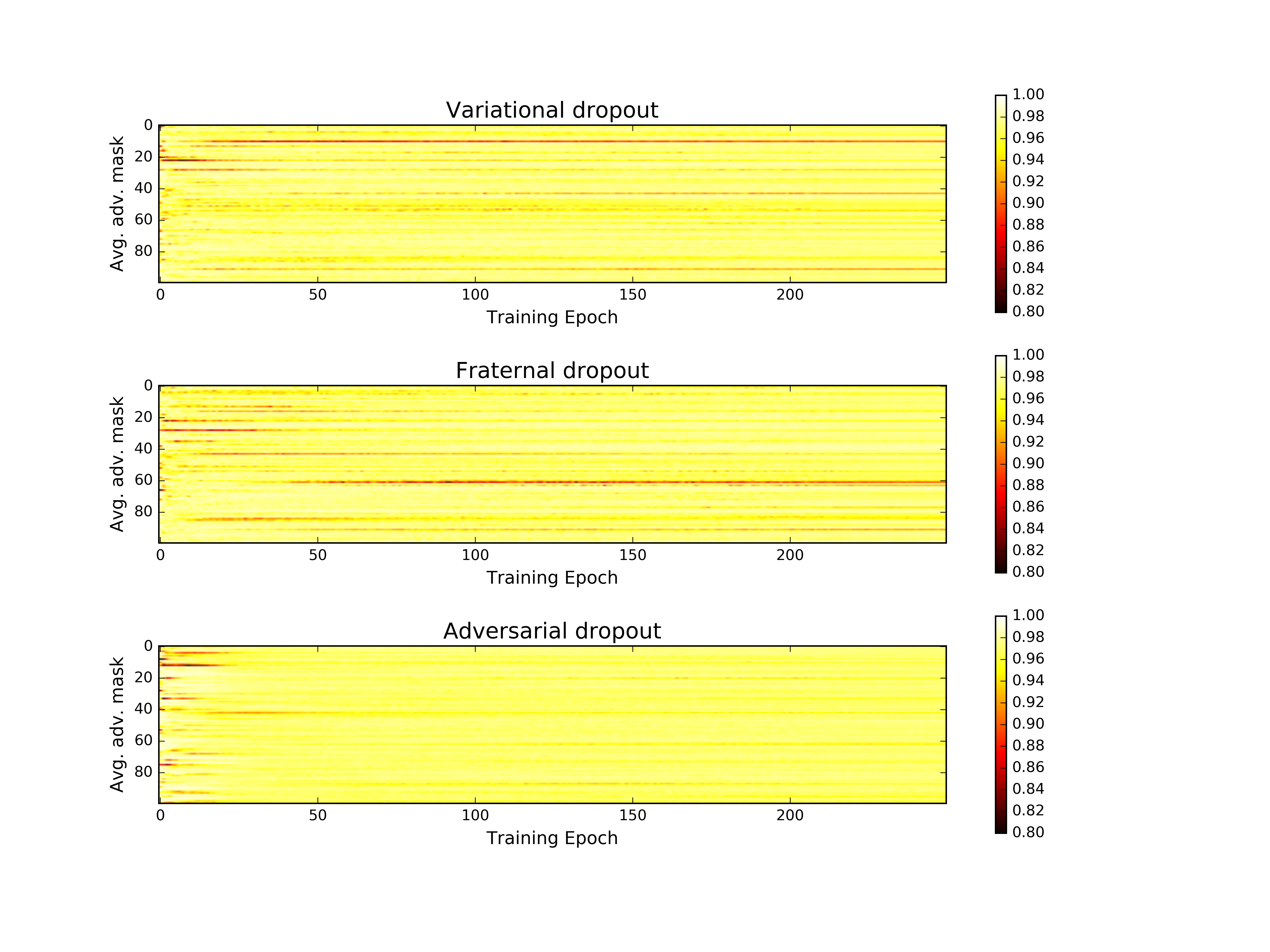}
\caption{Averages of the adversarial dropout masks over the test dataset through the training iterations.}
\end{figure}

Table 1 shows the test performances of the dropout-based regularizations. When applying variational dropout on the recurrent connections, the performance was improved from the unregularized LSTM. By adding regularization terms, the performance on the test dataset was improved in the following order: FD, and adversarial dropout ($K$=1, 2). 

In order to investigate how adversarial dropout training affects the accuracy distribution of the dropped subnetworks, we sampled random dropout masks and adversarial dropout masks respectively 500 times and tested the performance of the subnetwork reconfigured by the sampled dropout masks. Figure 3 (a) shows the histogram of the results. In this analysis, we applied the dropout probability, $p$=0.03, and the hyperparameters, $\delta$=0.03 and $K$=2, to the adversarial dropout. In the case of the variational dropout, we can see that the test performances of adversarially dropped networks were worse than the test performance of randomly dropped networks. By adding the regularization term of the FD, the performances of adversarially dropped networks were improved and caused the performance distribution of randomly dropped networks to move the right and its variance to reduce. That is, regularization of the FD indirectly improved the performance of the subnetwork in its worst case scenario. On the other hand, adversarial dropout directly controlled the sub-network in the worst case scenario. As a result, the performances of the base subnetwork and the adversarially perturbed networks were improved more than the performances in the case of FD regularization. 

\begin{table*}
\centering
\caption{Test error rates (\%) on the IMDB and Elec classification tasks. All models were pre-trained by neural language models. Our experiments were repeated five times.}
\begin{tabular}{clcc}
\toprule
&  Method  & IMDB & Elec \\
\midrule
\parbox[t]{1mm}{\multirow{5}{*}{\rotatebox[origin=c]{90}{\textbf{Reported}}}}
&One-hot CNN \cite{johnson2015semi,johnson2016supervised}	&6.05		&	5.87  \\
&One-hot bi-LSTM \cite{johnson2016supervised}	&5.94		&	5.55  \\
&Adversarial training  \cite{miyato2016adversarial}						&6.21		&	5.61	\\
&Virtual adversarial training 	\cite{miyato2016adversarial}				&5.91		&	5.54	\\
&Adversarial + virtual adversarial training \cite{miyato2016adversarial}			&6.09		&	\textbf{5.40}   \\
\midrule
\parbox[t]{1mm}{\multirow{14}{*}{\rotatebox[origin=c]{90}{\textbf{Our experiment}}}}
&\textbf{Our base model} & & \\
&LSTM 								& 7.164 ($\pm$0.070)		& 	6.155 ($\pm$0.092)	\\
\cmidrule{2-4} 
&\textbf{Input perturbation on embeddings} & & \\
&Adversarial training 							& 6.296 ($\pm$0.115)		& 	5.652 ($\pm$0.064)	\\
&Virtual adversarial training 						& 5.914 ($\pm$0.051)		& 	5.610 ($\pm$0.045)	\\
\cmidrule{2-4} 
&\textbf{Dropout perturbation on hidden states} & & \\
&Variational dropout 							& 7.042 ($\pm$0.059)		& 	6.114 ($\pm$0.074)	\\
&Expectation-Linearization dropout					& 6.358 ($\pm$0.065)		& 	5.734 ($\pm$0.052)	\\
&Fraternal dropout								& 6.349 ($\pm$0.060)		& 	5.785 ($\pm$0.085)	\\
&Adversarial dropout on the last hidden state   & 6.471 ($\pm$0.058)        & 	5.784 ($\pm$0.064)	\\
&Adversarial dropout on the recurrent connection ($K$=1)							& 6.155 ($\pm$0.053)		& 	5.622 ($\pm$0.051)	\\
&Adversarial dropout on the recurrent connection ($K$=2)							& 6.005 ($\pm$0.043)		& 	5.603 ($\pm$0.049)	\\
\cmidrule{2-4} 
&\textbf{Input and dropout perturbations} & & \\
&Adversarial dropout ($K$=1) + virtual adversarial training   		& 5.715 ($\pm$0.081)		& 	5.638 ($\pm$0.044)	\\
&Adversarial dropout ($K$=2) + virtual adversarial training   		& \textbf{5.687} ($\pm$0.068)		& 	5.621  ($\pm$0.058)	\\
\bottomrule
\end{tabular}
\end{table*}

We additionally investigated the adversarial dropout condition changes throughout the training phase. We calculated $\mathbb{E}_{\text{test}}[\boldsymbol{\epsilon^{\text{adv}}}]$ on every training epoch. Figure 3 (b) shows the visualized adversarial dropout masks. As can be seen, the adversarial dropout condition is equally spread over the test dataset in the case of adversarial dropout training, whereas the other cases show that a certain number of elements are selected more frequently for the adversarial dropout condition. These results show that adversarial dropout training stimulates the useless or the incorrectly learned neurons to better contribute to a more accurate prediction. 

\subsection{Semi-supervised text classification}

The text classification task is one of the most important tasks utilizing RNN architecture. We evaluated our method on two text datasets, IMDB and Elec. IMDB is a standard benchmark movie review dataset for sentiment classification\cite{maas2011learning}. Elec is a dataset on electronic product reviews from Amazon \cite{johnson2015semi}. Both datasets consist of labeled text data and unlabeled text data. In Elec, we removed some duplicated examples in labeled and unlabeled training dataset. We describe the detail summarization in the appendix.

Following previous studies on these datasets, we implemented a pre-training phase through neural language modeling and a training phase using the classification model. Most experiment settings are similar to \cite{miyato2016adversarial} (detail settings in the appendix).  Our implementation code will be available at \url{https://github.com/sungraepark/adversarial_dropout_text_classification}.

Table 2 shows the test performance on IMDB and Elec with each training method. The base LSTM achieved a 7.164\% error rate on IMDB and 6.155\% error rate on Elec with only embedding dropout. By adding variational dropout on the hidden states with the time-invariant dropout mask, the error rate was slightly reduced to 7.042\% and 6.114\%, respectively. When we tested EL regularizers, that are marked as the dropout perturbation in the table, the performance was improved in the following order: EL, FD, and adversarial dropout ($K$=1, 2). In the results, we found that adversarial dropout improved performance more than the FD. This might be caused by the fact that the regularization term of adversarial dropout is the upper bound of the regularization term of the FD. When comparing from adversarial dropout on the last hidden state, aka. original adversarial dropout, we found that applying adversarial dropout at the recurrent connections was more effective. Additionally, our iterative algorithm with $K$=2 provided the further performance gains. Furthermore, we evaluated a jointed method of adversarial dropout, which is dropout perturbation on hidden states, and virtual adversarial training, which is a linear perturbation on embedding space. As a result, we found that the jointed method achieved an impressive performance of 5.687\% on IMDB. 

\subsection{Word-level language models}

\begin{table*}
\centering
\caption{Perplexity on  word-level language modeling on Penn Treebank and WikiText-2.}
\resizebox{0.9\linewidth}{!}{
\begin{tabular}{lccccccc}
\toprule
&	\multicolumn{3}{c}{PTB} & & \multicolumn{3}{c}{WikiText-2} \\
\cmidrule{2-4} \cmidrule{6-8} 
  Model & Param. & Val. & Test && Param. & Val. & Test\\
\midrule
Variational dropout \cite{gal2016theoretically} 					&	51M    &71.1	&68.5   &&	28M    &91.5	&87.0 \\
5-layer RHN  \cite{melis2018on}					&	24M    &64.8	&62.2   &&	24M    &78.1	&75.6 \\
AWD-LSTM \cite{merity2017regularizing}		 			&	24M    &60.0	&58.3   &&	34M    &68.6	&65.8 \\
Fraternal dropout 	\cite{zolna2018fraternal}				&	24M    &58.9	&56.8   &&	34M    &66.8	&64.1 \\
\midrule
Adversarial dropout  				&	24M    & \textbf{58.7} $\pm$0.3	&\textbf{56.4} $\pm$0.2 
                                                       &&	34M    & \textbf{66.5}  $\pm$0.1	&\textbf{63.4} $\pm$0.2 \\
\bottomrule
\end{tabular}
}
\end{table*}

We investigated the performance of our regularization model in language modeling for two benchmark datasets, the Penn Treebank (PTB)\cite{mikolov2010recurrent} and WikiText-2 (WT2) dataset\cite{merity2017pointer}. We preprocessed both datasets: for PTB as specified by Mikolov et al. \cite{mikolov2010recurrent} and for WT2 as specified by Koehn et al. \cite{koehn2007moses}. In this experiment, we used AWD-LSTM 3-layer architecture that was introduced by Merity et al.\cite{merity2017regularizing}. Specifically, the architecture contains one 400 dimensional embedding layer and three LSTM layers whose dimensions are 1150 for first two layers and 400 for the third layer. The embedding matrix and the weight matrix of the last layer for predictions were tied \cite{merity2017revisiting}. In the training phase, Merity et al.\cite{merity2017regularizing} applied DropConnect to improve model performances. In this experiment, we left a vast majority of hyperparameters used in the baseline model, i.e. embedding and hidden sizes, learning rate, and so on. Instead, we controlled the hyperparameters, $\delta$ and $K$, related to the adversarial dropout. We did not use the additional regularization terms except for the L2 norm for weight parameters. Our implementation code will be available at \url{https://github.com/sungraepark/adversarial_dropout_lm}.

Previous results with LSTM showed that fine-tuning is important to achieve state-of-the-art performances \cite{merity2017regularizing,li2018towards}, so we fine-tuned model parameters once without regularization after learning was over with our regularization term. Table 3 shows the perplexity on both the PTB and WikiText-2 validation and test datasets. Our approach showed an advanced performance compared to existing benchmarks.  

\section{Conclusion}

The improvement of generalization performance in RNNs is required for the successful application of sequential data processing. The existing methods utilizing the dropout depend on the random dropout mask without considering a guided sampling on the dropout mask. In contrast, we developed an RNN version of adversarial dropout, which is a deterministic dropout technique to find a subnetwork inferior to target predictions. Specifically, we proved that the regularizer with the adversarial dropout is the upper bound of the EL and FD regularizers. Additionally, we found a way to measure which recurrent neurons are mainly used for target predictions. Furthermore, we improved the algorithm to find the adversarial dropout condition. In our experiments, we showed that the adversarial dropout for RNNs improved generalization performance on the sequential MNIST tasks, the semi-supervised text classification tasks, and word-level language modeling tasks. More importantly, we achieved a highly advanced performance of 5.687\% on IMDB when applying an adversarial perturbation on word embeddings and an adversarial dropout perturbation, together.  

\section{Acknowledgement}
This research was supported by Basic Science Research Program through the National Research Foundation of Korea(NRF) funded by the Ministry of Education(NRF-2018R1C1B6008652)

\bibliographystyle{aaai}
\bibliography{reff}

\begin{thebibliography}{}

\bibitem[\protect\citeauthoryear{Bachman, Alsharif, and
  Precup}{2015}]{bachman2014learning}
Bachman, P.; Alsharif, O.; and Precup, D.
\newblock 2015.
\newblock Learning with pseudo-ensembles.
\newblock In {\em Advances in Neural Information Processing Systems},
  3365--3373.

\bibitem[\protect\citeauthoryear{Bengio \bgroup et al\mbox.\egroup
  }{2003}]{bengio2003neural}
Bengio, Y.; Ducharme, R.; Vincent, P.; and Jauvin, C.
\newblock 2003.
\newblock A neural probabilistic language model.
\newblock {\em Journal of machine learning research} 3(Feb):1137--1155.

\bibitem[\protect\citeauthoryear{Bishop}{1995}]{bishop1995training}
Bishop, C.~M.
\newblock 1995.
\newblock Training with noise is equivalent to tikhonov regularization.
\newblock {\em Neural computation} 7(1):108--116.

\bibitem[\protect\citeauthoryear{Bul{\`o}, Porzi, and
  Kontschieder}{2016}]{bulo2016dropout}
Bul{\`o}, S.~R.; Porzi, L.; and Kontschieder, P.
\newblock 2016.
\newblock Dropout distillation.
\newblock In {\em International Conference on Machine Learning},  99--107.

\bibitem[\protect\citeauthoryear{Gal and
  Ghahramani}{2016}]{gal2016theoretically}
Gal, Y., and Ghahramani, Z.
\newblock 2016.
\newblock A theoretically grounded application of dropout in recurrent neural
  networks.
\newblock In {\em Advances in Neural Information Processing Systems},
  1019--1027.

\bibitem[\protect\citeauthoryear{Goodfellow, Shlens, and
  Szegedy}{2015}]{goodfellow2014explaining}
Goodfellow, I.; Shlens, J.; and Szegedy, C.
\newblock 2015.
\newblock Explaining and harnessing adversarial examples.
\newblock In {\em International Conference on Learning Representations}.

\bibitem[\protect\citeauthoryear{Ioffe and Szegedy}{2015}]{ioffe2015batch}
Ioffe, S., and Szegedy, C.
\newblock 2015.
\newblock Batch normalization: Accelerating deep network training by reducing
  internal covariate shift.
\newblock In {\em International conference on machine learning},  448--456.

\bibitem[\protect\citeauthoryear{Johnson and Zhang}{2015}]{johnson2015semi}
Johnson, R., and Zhang, T.
\newblock 2015.
\newblock Semi-supervised convolutional neural networks for text categorization
  via region embedding.
\newblock In {\em Advances in neural information processing systems},
  919--927.

\bibitem[\protect\citeauthoryear{Johnson and
  Zhang}{2016}]{johnson2016supervised}
Johnson, R., and Zhang, T.
\newblock 2016.
\newblock Supervised and semi-supervised text categorization using lstm for
  region embeddings.
\newblock In {\em International Conference on Machine Learning},  526--534.

\bibitem[\protect\citeauthoryear{Kingma and Ba}{2014}]{kingma2014adam}
Kingma, D., and Ba, J.
\newblock 2014.
\newblock Adam: A method for stochastic optimization.
\newblock {\em arXiv preprint arXiv:1412.6980}.

\bibitem[\protect\citeauthoryear{Kingma, Salimans, and
  Welling}{2015}]{kingma2015variational}
Kingma, D.~P.; Salimans, T.; and Welling, M.
\newblock 2015.
\newblock Variational dropout and the local reparameterization trick.
\newblock In {\em Advances in Neural Information Processing Systems},
  2575--2583.

\bibitem[\protect\citeauthoryear{Koehn \bgroup et al\mbox.\egroup
  }{2007}]{koehn2007moses}
Koehn, P.; Hoang, H.; Birch, A.; Callison-Burch, C.; Federico, M.; Bertoldi,
  N.; Cowan, B.; Shen, W.; Moran, C.; Zens, R.; et~al.
\newblock 2007.
\newblock Moses: Open source toolkit for statistical machine translation.
\newblock In {\em Proceedings of the 45th annual meeting of the ACL on
  interactive poster and demonstration sessions},  177--180.
\newblock Association for Computational Linguistics.

\bibitem[\protect\citeauthoryear{Laine and Aila}{2017}]{Laine2017iclr}
Laine, S., and Aila, T.
\newblock 2017.
\newblock Temporal ensembling for semi-supervised learning.
\newblock In {\em Proc. International Conference on Learning Representations
  (ICLR)}.

\bibitem[\protect\citeauthoryear{Li \bgroup et al\mbox.\egroup
  }{2018}]{li2018towards}
Li, Z.; He, D.; Tian, F.; Chen, W.; Qin, T.; Wang, L.; and Liu, T.-Y.
\newblock 2018.
\newblock Towards binary-valued gates for robust lstm training.
\newblock In {\em International Conference on Machine Learning}.

\bibitem[\protect\citeauthoryear{Ma \bgroup et al\mbox.\egroup
  }{2017}]{ma2016dropout}
Ma, X.; Gao, Y.; Hu, Z.; Yu, Y.; Deng, Y.; and Hovy, E.
\newblock 2017.
\newblock Dropout with expectation-linear regularization.
\newblock In {\em International Conference on Learning Representations}.

\bibitem[\protect\citeauthoryear{Maas \bgroup et al\mbox.\egroup
  }{2011}]{maas2011learning}
Maas, A.~L.; Daly, R.~E.; Pham, P.~T.; Huang, D.; Ng, A.~Y.; and Potts, C.
\newblock 2011.
\newblock Learning word vectors for sentiment analysis.
\newblock In {\em Proceedings of the 49th annual meeting of the association for
  computational linguistics: Human language technologies-volume 1},  142--150.
\newblock Association for Computational Linguistics.

\bibitem[\protect\citeauthoryear{Melis, Dyer, and Blunsom}{2018}]{melis2018on}
Melis, G.; Dyer, C.; and Blunsom, P.
\newblock 2018.
\newblock On the state of the art of evaluation in neural language models.
\newblock In {\em International Conference on Learning Representations}.

\bibitem[\protect\citeauthoryear{Merity \bgroup et al\mbox.\egroup
  }{2017}]{merity2017pointer}
Merity, S.; Xiong, C.; Bradbury, J.; and Socher, R.
\newblock 2017.
\newblock Pointer sentinel mixture models.
\newblock In {\em International Conference on Learning Representations}.

\bibitem[\protect\citeauthoryear{Merity, Keskar, and
  Socher}{2018}]{merity2017regularizing}
Merity, S.; Keskar, N.~S.; and Socher, R.
\newblock 2018.
\newblock Regularizing and optimizing lstm language models.
\newblock In {\em International Conference on Learning Representations}.

\bibitem[\protect\citeauthoryear{Merity, McCann, and
  Socher}{2017}]{merity2017revisiting}
Merity, S.; McCann, B.; and Socher, R.
\newblock 2017.
\newblock Revisiting activation regularization for language rnns.
\newblock {\em arXiv preprint arXiv:1708.01009}.

\bibitem[\protect\citeauthoryear{Mikolov \bgroup et al\mbox.\egroup
  }{2010}]{mikolov2010recurrent}
Mikolov, T.; Karafi{\'a}t, M.; Burget, L.; {\v{C}}ernock{\`y}, J.; and
  Khudanpur, S.
\newblock 2010.
\newblock Recurrent neural network based language model.
\newblock In {\em Eleventh Annual Conference of the International Speech
  Communication Association}.

\bibitem[\protect\citeauthoryear{Miyato \bgroup et al\mbox.\egroup
  }{2017}]{miyato2017virtual}
Miyato, T.; Maeda, S.-i.; Koyama, M.; and Ishii, S.
\newblock 2017.
\newblock Virtual adversarial training: a regularization method for supervised
  and semi-supervised learning.
\newblock {\em arXiv preprint arXiv:1704.03976}.

\bibitem[\protect\citeauthoryear{Miyato, Dai, and
  Goodfellow}{2016}]{miyato2016adversarial}
Miyato, T.; Dai, A.~M.; and Goodfellow, I.
\newblock 2016.
\newblock Adversarial training methods for semi-supervised text classification.
\newblock {\em arXiv preprint arXiv:1605.07725}.

\bibitem[\protect\citeauthoryear{Miyato, Dai, and
  Goodfellow}{2017}]{miyato2017adversarial}
Miyato, T.; Dai, A.~M.; and Goodfellow, I.
\newblock 2017.
\newblock Adversarial training methods for semi-supervised text classification.

\bibitem[\protect\citeauthoryear{Moon \bgroup et al\mbox.\egroup
  }{2015}]{moon2015rnndrop}
Moon, T.; Choi, H.; Lee, H.; and Song, I.
\newblock 2015.
\newblock Rnndrop: A novel dropout for rnns in asr.
\newblock In {\em Automatic Speech Recognition and Understanding (ASRU), 2015
  IEEE Workshop on},  65--70.
\newblock IEEE.

\bibitem[\protect\citeauthoryear{Papernot \bgroup et al\mbox.\egroup
  }{2016}]{papernot2016limitations}
Papernot, N.; McDaniel, P.; Jha, S.; Fredrikson, M.; Celik, Z.~B.; and Swami,
  A.
\newblock 2016.
\newblock The limitations of deep learning in adversarial settings.
\newblock In {\em Security and Privacy (EuroS\&P), 2016 IEEE European Symposium
  on},  372--387.
\newblock IEEE.

\bibitem[\protect\citeauthoryear{Park \bgroup et al\mbox.\egroup
  }{2018}]{park2017adversarial}
Park, S.; Park, J.-K.; Shin, S.-J.; and Moon, I.-C.
\newblock 2018.
\newblock Adversarial dropout for supervised and semi-supervised learning.
\newblock In {\em AAAI}, volume~4, ~12.

\bibitem[\protect\citeauthoryear{Salimans and
  Kingma}{2016}]{salimans2016weight}
Salimans, T., and Kingma, D.~P.
\newblock 2016.
\newblock Weight normalization: A simple reparameterization to accelerate
  training of deep neural networks.
\newblock In {\em Advances in Neural Information Processing Systems},
  901--901.

\bibitem[\protect\citeauthoryear{Samanta and Mehta}{2017}]{samanta2017towards}
Samanta, S., and Mehta, S.
\newblock 2017.
\newblock Towards crafting text adversarial samples.
\newblock {\em arXiv preprint arXiv:1707.02812}.

\bibitem[\protect\citeauthoryear{Semeniuta, Severyn, and
  Barth}{2016}]{Semeniuta45423}
Semeniuta, S.; Severyn, A.; and Barth, E.
\newblock 2016.
\newblock Recurrent dropout without memory loss.
\newblock Technical report.

\bibitem[\protect\citeauthoryear{Srivastava \bgroup et al\mbox.\egroup
  }{2014}]{srivastava2014dropout}
Srivastava, N.; Hinton, G.~E.; Krizhevsky, A.; Sutskever, I.; and
  Salakhutdinov, R.
\newblock 2014.
\newblock Dropout: a simple way to prevent neural networks from overfitting.
\newblock {\em Journal of Machine Learning Research} 15(1):1929--1958.

\bibitem[\protect\citeauthoryear{Tarvainen and
  Valpola}{2017}]{tarvainen2017mean}
Tarvainen, A., and Valpola, H.
\newblock 2017.
\newblock Mean teachers are better role models: Weight-averaged consistency
  targets improve semi-supervised deep learning results.
\newblock In {\em Advances in neural information processing systems},
  1195--1204.

\bibitem[\protect\citeauthoryear{Tieleman and
  Hinton}{2012}]{tieleman2012lecture}
Tieleman, T., and Hinton, G.
\newblock 2012.
\newblock Lecture 6.5-rmsprop: Divide the gradient by a running average of its
  recent magnitude.
\newblock {\em COURSERA: Neural networks for machine learning} 4(2):26--31.

\bibitem[\protect\citeauthoryear{Wan \bgroup et al\mbox.\egroup
  }{2013}]{wan2013regularization}
Wan, L.; Zeiler, M.; Zhang, S.; Le~Cun, Y.; and Fergus, R.
\newblock 2013.
\newblock Regularization of neural networks using dropconnect.
\newblock In {\em International Conference on Machine Learning},  1058--1066.

\bibitem[\protect\citeauthoryear{Zaremba, Sutskever, and
  Vinyals}{2015}]{zaremba2014recurrent}
Zaremba, W.; Sutskever, I.; and Vinyals, O.
\newblock 2015.
\newblock Recurrent neural network regularization.
\newblock {\em ICLR}.

\bibitem[\protect\citeauthoryear{Zolna \bgroup et al\mbox.\egroup
  }{2018}]{zolna2018fraternal}
Zolna, K.; Arpit, D.; Suhubdy, D.; and Bengio, Y.
\newblock 2018.
\newblock Fraternal dropout.
\newblock In {\em International Conference on Learning Representations}.

\end{thebibliography}

\section{Appendix 1. Proofs of remark 1}

\textbf{Remark 1.} Let $\boldsymbol{\epsilon}$ be a i.i.d. dropout mask and $\boldsymbol{p}_t(\boldsymbol{x},\boldsymbol{\epsilon} ; \boldsymbol{\theta} ) \in \mathbb{R}^M$ be the prediction at time $t$ where $M$ is a size of the ouput dimension. For a simple description, we denote $p_{t, i}(\boldsymbol{\epsilon})$ be a $i^{\text{th}}$ element of $\boldsymbol{p}_t(\boldsymbol{x},\boldsymbol{\epsilon} ; \boldsymbol{\theta} )$. Then,
\begin{gather}
\begin{split}
\mathrm{E}_{\boldsymbol{\epsilon}}[\mathcal{R}_t^{\text{AdD}}(\boldsymbol{x}, \boldsymbol{\epsilon} ;\theta)] &=  
  \sum_i \mathrm{V}_{\boldsymbol{\epsilon}}[p_{t, i}(\boldsymbol{\epsilon})] + \sum_i \mathrm{V}_{\boldsymbol{\epsilon}}[p_{t, i}(\boldsymbol{\epsilon}^{\text{adv}})] \\
& -2  \sum_i \mathrm{Cov}_{\boldsymbol{\epsilon}}[p_{t, i}(\boldsymbol{\epsilon}), p_{t, i}(\boldsymbol{\epsilon}^{\text{adv}})] \\
& \quad +  \sum_i \big(\mathrm{E}_{\boldsymbol{\epsilon}}[p_{t, i}(\boldsymbol{\epsilon})]- \mathrm{E}_{\boldsymbol{\epsilon}}[p_{t, i}(\boldsymbol{\epsilon}^{\text{adv}})]\big)^2 .
\end{split}
\end{gather}
$Proof.$ 
\begin{gather}
\begin{split}
\mathrm{E}_{\boldsymbol{\epsilon}}[\mathcal{R}_t^{\text{AdD}}&(\boldsymbol{x}, \boldsymbol{\epsilon} ;\theta)]  
= \mathrm{E}_{\boldsymbol{\epsilon}}[ || \boldsymbol{p}_t(\boldsymbol{x},\boldsymbol{\epsilon} ; \boldsymbol{\theta} ) - \boldsymbol{p}_t(\boldsymbol{x},\boldsymbol{\epsilon}^{\text{adv}} ; \boldsymbol{\theta} ) ||^2_2 ] \\
& = \mathrm{E}_{\boldsymbol{\epsilon}} \begin{bmatrix} || \boldsymbol{p}_t(\boldsymbol{x},\boldsymbol{\epsilon} ; \boldsymbol{\theta} )||^2_2 + ||\boldsymbol{p}_t(\boldsymbol{x},\boldsymbol{\epsilon}^{\text{adv}} ; \boldsymbol{\theta} ) ||^2_2  \\ 
   - 2\boldsymbol{p}_t(\boldsymbol{x},\boldsymbol{\epsilon} ; \boldsymbol{\theta} )^{T}\boldsymbol{p}_t(\boldsymbol{x},\boldsymbol{\epsilon}^{\text{adv}} ; \boldsymbol{\theta} ) \end{bmatrix} \\
& = \sum_{i=1}^{M} \big\{ \mathrm{E}_{\boldsymbol{\epsilon}}[p_{t, i}(\boldsymbol{\epsilon})^2]+\mathrm{E}_{\boldsymbol{\epsilon}}[p_{t, i}(\boldsymbol{\epsilon}^{\text{adv}})^2] \\[-2ex]
& \quad \quad \quad - 2\mathrm{E}_{\boldsymbol{\epsilon}}[p_{t, i}(\boldsymbol{\epsilon})p_{t, i}(\boldsymbol{\epsilon}^{\text{adv}})] \big\} \\
& = \sum_{i=1}^{M} \big\{ \mathrm{V}_{\boldsymbol{\epsilon}}[p_{t, i}(\boldsymbol{\epsilon})]+\mathrm{V}_{\boldsymbol{\epsilon}}[p_{t, i}(\boldsymbol{\epsilon}^{\text{adv}})] + \mathrm{E}^2_{\boldsymbol{\epsilon}}[p_{t, i}(\boldsymbol{\epsilon})] \\[-2ex]
& \quad \quad \quad +\mathrm{E}^2_{\boldsymbol{\epsilon}}[p_{t, i}(\boldsymbol{\epsilon}^{\text{adv}})]  -2\mathrm{E}_{\boldsymbol{\epsilon}}[p_{t, i}(\boldsymbol{\epsilon})p_{t, i}(\boldsymbol{\epsilon}^{\text{adv}}) \big\} \\
& = \sum_{i=1}^{M} \big\{ \mathrm{V}_{\boldsymbol{\epsilon}}[p_{t, i}(\boldsymbol{\epsilon})]+\mathrm{V}_{\boldsymbol{\epsilon}}[p_{t, i}(\boldsymbol{\epsilon}^{\text{adv}})]  \\[-2ex] 
& \quad \quad \quad + \big(\mathrm{E}_{\boldsymbol{\epsilon}}[p_{t, i}(\boldsymbol{\epsilon})]- \mathrm{E}_{\boldsymbol{\epsilon}}[p_{t, i}(\boldsymbol{\epsilon}^{\text{adv}})]\big)^2 \\
& \qquad \quad  -2\mathrm{E}_{\boldsymbol{\epsilon}}[p_{t, i}(\boldsymbol{\epsilon})p_{t, i}(\boldsymbol{\epsilon}^{\text{adv}})] \\
& \qquad \quad +2 \mathrm{E}_{\boldsymbol{\epsilon}}[p_{t, i}(\boldsymbol{\epsilon})]\mathrm{E}_{\boldsymbol{\epsilon}}[p_{t, i}(\boldsymbol{\epsilon}^{\text{adv}})] \big\} \\
& = \sum_{i=1}^{M} \big\{ \mathrm{V}_{\boldsymbol{\epsilon}}[p_{t, i}(\boldsymbol{\epsilon})]+\mathrm{V}_{\boldsymbol{\epsilon}}[p_{t, i}(\boldsymbol{\epsilon}^{\text{adv}})]  \\[-2ex]
& \qquad \quad + \big(\mathrm{E}_{\boldsymbol{\epsilon}}[p_{t, i}(\boldsymbol{\epsilon})]- \mathrm{E}_{\boldsymbol{\epsilon}}[p_{t, i}(\boldsymbol{\epsilon}^{\text{adv}})]\big)^2 \\
& \qquad \quad -2\mathrm{Cov}_{\boldsymbol{\epsilon}}[p_{t, i}(\boldsymbol{\epsilon}), p_{t, i}(\boldsymbol{\epsilon}^{\text{adv}})] \big\} \\
\end{split}
\end{gather}

\section{Appendix 2. Proofs of proposition 1}

\textbf{Proposition 1.}  $\mathcal{R}^{\text{EL}}(\boldsymbol{x};\theta) \leq  \mathcal{R}^{\text{AdD}}(\boldsymbol{x}, \mathrm{E}_{\boldsymbol{\epsilon}}[\boldsymbol{\epsilon}] ;\theta)$ when $\boldsymbol{\epsilon} \in \{ \boldsymbol{\epsilon} | \| E_{\boldsymbol{\epsilon}}[\boldsymbol{\epsilon}] - \boldsymbol{\epsilon} \|_2 \leq \delta\} $.

$Proof$. 
\begin{gather}
\begin{split}
\mathcal{R}^{\text{AdD}}&(\boldsymbol{x}, E_{\boldsymbol{\epsilon}}[\boldsymbol{\epsilon}] ;\theta) = \sum_{t=1}^{T} \lambda_t \mathrm{D}\big[\boldsymbol{p}_t(\boldsymbol{x},E_{\boldsymbol{\epsilon}}[\boldsymbol{\epsilon}] ; \boldsymbol{\theta} ) || \boldsymbol{p}_t(\boldsymbol{x},\boldsymbol{\epsilon}^{\text{adv}} ; \boldsymbol{\theta} ) \big] \\
& = \lim_{S \to\infty} \frac{1}{S} \sum_{s=1}^S \sum_{t=1}^{T} \lambda_t \mathrm{D}\big[\boldsymbol{p}_t(\boldsymbol{x},E_{\boldsymbol{\epsilon}}[\boldsymbol{\epsilon}] ; \boldsymbol{\theta} ) || \boldsymbol{p}_t(\boldsymbol{x},\boldsymbol{\epsilon}^{\text{adv}} ; \boldsymbol{\theta} ) \big] \\
& \geq \lim_{S \to\infty} \frac{1}{S} \sum_{s=1}^S \sum_{t=1}^{T} \lambda_t \mathrm{D}\big[\boldsymbol{p}_t(\boldsymbol{x},E_{\boldsymbol{\epsilon}}[\boldsymbol{\epsilon}] ; \boldsymbol{\theta} ) || \boldsymbol{p}_t(\boldsymbol{x},\boldsymbol{\epsilon}^{s} ; \boldsymbol{\theta} ) \big] \\
& =  \mathrm{E}_{\boldsymbol{\epsilon}}\Big[ \sum_{t=1}^{T} \lambda_t \mathrm{D}\big[\boldsymbol{p}_t(\boldsymbol{x},\mathrm{E}_{\boldsymbol{\epsilon}}[\boldsymbol{\epsilon}] ; \boldsymbol{\theta} ) || \boldsymbol{p}_t(\boldsymbol{x},\boldsymbol{\epsilon} ; \boldsymbol{\theta} ) \big] \Big] \\
& = \mathcal{R}^{\text{EL}}(\boldsymbol{x};\theta), \\
\end{split}
\end{gather}
where $\boldsymbol{\epsilon}^{s}$ is a sampled dropout mask, which domain is $\boldsymbol{\epsilon} \in \{ \boldsymbol{\epsilon} | \| E_{\boldsymbol{\epsilon}}[\boldsymbol{\epsilon}] - \boldsymbol{\epsilon} \|_2 \leq \delta\} $.

\section{Appendix 3. A greedy algorithm to find the adversarial dropout mask}

\begin{algorithm}[h]
\caption{Filp function for adversarial dropout conditions}\label{fast_algo}
\SetKwInOut{Input}{Input}
\SetKwInOut{Output}{Output}

\Input{$\boldsymbol{\epsilon}^{0}$ is an base dropout mask for the constraint}
\Input{$\boldsymbol{\epsilon}^{s}$ is an initial dropout mask}
\Input{$\delta$ is a hyper-parameter for the boundary}
\Input{$\mathbf{IM}$ is the influence map}
\Input{$H$ is the layer dimension.}
\Output{$\boldsymbol{\epsilon}_{adv}$}
\Begin{
	$\widetilde{\mathbf{IM}} \longleftarrow (1-2\boldsymbol{\epsilon}^{s}) \odot \mathbf{IM}$ \\
	$\boldsymbol{i} \longleftarrow $ Arg Sort $\widetilde{\mathbf{IM}}$ as $\widetilde{\mathbf{IM}}_{i_1} \geq ... \geq \widetilde{\mathbf{IM}}_{i_H}$  \\
	$\boldsymbol{\epsilon}^{adv} \longleftarrow \boldsymbol{\epsilon}^{s}$ \\
	$d \longleftarrow 1$ \\
	\While{$\| \boldsymbol{\epsilon}^0 - \boldsymbol{\epsilon}^{adv} \|_2 \leq \delta H$ and $d \leq H$}{
		\uIf{$\epsilon^{adv}_{i_d}=0$} {
			$\epsilon^{adv}_{i_d} \longleftarrow 1$ \# flipping
		}
		\ElseIf{$\epsilon^{adv}_{i_d}=1$} {
			$\epsilon^{adv}_{i_d} \longleftarrow 0$ \# flipping
		}
		$d \longleftarrow d + 1$
	}
}
\end{algorithm}
Algorithm 1 shows the flip algorithm that captures an adversarial dropout condition utilizing calculated influence map.

\section{Appendix 4. Experiment settings for sequential MNIST tasks}

The baseline model included a single LSTM layer of 100 units with a softmax classifier to produce a prediction from the final hidden state. We applied the variational dropout, which deactivates the hidden state with the time-invariant dropout. All networks are trained using RMSProp \cite{tieleman2012lecture} with a decay rate of 0.5 for the moving average of gradient norms. We annealed the learning rate to zero during the last 50 epoch with the initial learning rate as 0.001. The gradients were clipped to a maximum norm of 1. For the variational dropout and FD, we tested dropout probabilities and we reported the best case of the performances (when $p$=0.1). For the DE regularizers, we set $\lambda_t=0$ in all time step except for the last time step with $\lambda_t=1$ because the final hidden states are only used for predictions.Additionally, we used Jensen-Shannon (JS) divergence for the distance metric of the two perturbed networks. For the adversarial dropout, we use the fully connected RNN as the base network. In other words, we set $\boldsymbol{\epsilon^{0}}=\mathrm{E}_{\boldsymbol{\epsilon}}[\boldsymbol{\epsilon}]$. For the constraint of the adversarial dropout, we set $\delta=0.03$, which represents the maximum changes from the base dropout mask as 3\%. For the initial dropout mask of our proposed algorithm, we set a dropout mask, whose randomly selected one element is flipped from the base network, as the initial dropout mask. Our implementation code is available at \url{https://github.com/sungraepark/AdvDrop-sMNIST}. 

\section{Appendix 5. Experiment settings for semi-supervised text classification tasks}

We initialized the word embedding matrix and LSTM weights with a pre-trained recurrent language model \cite{bengio2003neural} that was trained on both labeled and unlabeled examples. We set the word embedding dimension $D$ as 256 on IMDB and 512 on Elec and adopted a single-layer LSTM with 1024 hidden units. For the construction of the loss, we used a sampled softmax loss with 1024 candidate samples for training phase. The model was optimized by the Adam optimizer \cite{kingma2014adam} with momentum parameters of $\beta_1=0.9$ and $\beta_2=0.999$. The learning rate was initially set as 0.001 and was exponentially decayed with the decay rate of 0.9999 at each training step. The batch size was 256 and the gradient was clipped with norm set to 1.0 on all the parameters except word embeddings. We trained for 100,000 steps. For regularization, we applied the variational dropout on the word embedding layer with 0.5 dropout rate. The pre-training with the recurrent language model was very effective on classification performance on all the datasets. We didn't try to regularize this language models with the adversarial dropout for the fair comparison with the other baselines. However, in this pre-training step, the adversarial dropout can be applied.

After pre-training, we trained the text classification model with adversarial dropout. Between the softmax layer for the target $y$ and the final output of the LSTM, we added a hidden layer with 30 units. The activation function on the hidden layer was ReLU. For optimization, we again used the Adam optimizer, with 0.0005 initial learning rate 0.9998 exponential decay. Batch sizes are 64 for calculating the loss of the negative log-likelihood and the regularization term of adversarial dropout. We iterated 15,000 training steps for IMDB and 10,000 training steps for Elec. We again applied gradient clipping with the norm as 1.0 on all the parameters except the word embedding. We also used truncated backpropagation up to 400 words, and also generated the regularization terms. All models contain the dropout layer on the word-embeddings. We note that most of the experiment settings are same with the experimental settings specified by Miyato et al. \cite{miyato2017adversarial}. For the DE regularizers, we set $\lambda_t=0$ in all time step except for the last time step with $\lambda_t=1$ because the text class prediction is performed at the final time-step. Additionally, we used Jensen-Shannon (JS) divergence for the distance metric of the two perturbed networks. For the hyperparameters for the baseline models, we conducted the validation phase to identify the hyperparameters for the variational dropout and the FD regularization. For the hyperparameters of the adversarial dropout, we chose $\boldsymbol{\epsilon^{0}}=\mathrm{E}_{\boldsymbol{\epsilon}}[\boldsymbol{\epsilon}]$, which indicates the fully connected RNN, and set $\delta=0.04$, which represents the maximum changes from the base dropout mask as 4\%. We repeated training and evaluation five times with different random seeds. For the initial dropout mask of our proposed algorithm, we used $\mathrm{E}_{\boldsymbol{\epsilon}}[\boldsymbol{\epsilon}]$ as the initial dropout mask. We should note that there is another stochastic process on the word-embedding layer, so the zero gradient does not happen in the case. 

Our implementation code is available at \url{https://github.com/sungraepark/adversarial_dropout_text_classification}. 

\section{Appendix 6. Experiment settings for word-level language modeling tasks}

Our experiments on the word-level language modeling tasks were conducted based on the implementations as specified by Zolna et al. \cite{zolna2018fraternal}. Therefore, most hyperparameters are same with the experiments of \cite{zolna2018fraternal}. For the adversarial dropout, we only controlled dropout masks on the recurrent transition. For the hyperparameters of the adversarial dropout, we set $\boldsymbol{\epsilon^{0}}=\mathrm{E}_{\boldsymbol{\epsilon}}[\boldsymbol{\epsilon}]$ and $\delta=0.06$, which indicates that the maximum change rate was 6\%. For the initial dropout mask of our proposed algorithm, we used $\mathrm{E}_{\boldsymbol{\epsilon}}[\boldsymbol{\epsilon}]$ as the initial dropout mask. We should note that there is another stochastic process on the word-embedding layer, so the zero gradient does not happen in the case. 

\end{document}